\documentclass{article}
\usepackage{ijcai22}

\usepackage{times}
\usepackage{soul}
\usepackage{url}
\usepackage[hidelinks]{hyperref}
\usepackage[utf8]{inputenc}
\usepackage[small]{caption}
\usepackage{graphicx}
\usepackage{booktabs}
\usepackage{algorithm}
\usepackage{microtype}
\usepackage{makecell}
\usepackage{pifont}
\newcommand{\cmark}{\ding{51}}%
\newcommand{\xmark}{\ding{55}}%
\usepackage[table]{xcolor}
\definecolor{c1}{HTML}{FF6600}
\definecolor{c2}{HTML}{0c8918}
\definecolor{c3}{HTML}{b35c44}
\definecolor{c4}{HTML}{ff0097}
\definecolor{c5}{HTML}{6666FF}
\definecolor{c6}{HTML}{FFCC33}
\usepackage{color}

\usepackage{arydshln}
\usepackage{multirow}
\usepackage{algpseudocode}  
\usepackage{amsmath,amssymb,amsfonts}
\usepackage{subfigure}
\usepackage{CJK}
\usepackage{marvosym}
\usepackage{amsthm}
\usepackage{aligned-overset}

\title{Sem4SAP: Synonymous Expression Mining From Open Knowledge Graph For Language Model Synonym-Aware Pretraining }

\author{
Zhouhong Gu\thanks{Equal Contribution}$^{,1}$
\and
Sihang Jiang$^{*,1}$\and
Wenhao Huang$^{1}$ \and
Jiaqing Liang$^2$ \and \\ 
Hongwei Feng\thanks{Corresponding Authors}$^{,1}$\and
Yanghua Xiao$^{\dagger,1,3}$\and
\affiliations
$^1$Shanghai Key Laboratory of Data Science, School of Computer Science, Fudan University, China\\
$^2$School of Data Science, Fudan University\\
$^3$Fudan-Aishu Cognitive Intelligence Joint Research Center
\emails
\{zhgu20, liangjiaqing, shawyh\}@fudan.edu.cn,
tedsihangjiang@gmail.com,
whhuang21@m.fudan.edu.cn,
}

\begin{document}

\begin{CJK}{UTF8}{gbsn}
\maketitle

\begin{abstract}
    The model's ability to understand synonymous expression is crucial in many kinds of downstream tasks.
It will make the model to better understand the similarity between context, and more robust to the synonym substitution attack.
However, many Pretrained Language Model (PLM) lack synonym knowledge due to limitation of small-scale synsets and PLM's pretraining objectives.
In this paper, we propose a framework called Sem4SAP to mine synsets from Open Knowledge Graph (Open-KG) and using the mined synsets to do synonym-aware pretraining for language models.
We propose to coarsly filter the content in Open-KG and use the frequency information to better help the clustering process under low-resource unsupervised conditions.
We expand the mined synsets by migrating core semantics between synonymous expressions.
We also propose two novel and effective synonym-aware pre-training methods for injecting synonym knowledge into PLMs.
Extensive experiments demonstrate that Sem4SAP can dramatically outperform the original PLMs and other baselines on ten different tasks.

\end{abstract}

\section{Introduction}
\label{01}
Synonymous expressions are words, morphemes, or phrases that mean exactly or nearly the same to each other~\cite{stanojevic2009cognitive}.
In the field of Natural Language Process~(NLP), synonymous expressions play an important role in various applications.
For example, many downstream tasks have suffered from synonym substitution attack~\cite{chiang2022far}, the attack brings in imperceptible perturbations,
while not changing human predictions, will make a well-trained NLP model behave much worse than usual.
And synonym knowledge facilitates models to capture fine-grained semantic relations in some similarity-oriented tasks in the field of Information Retrieval~(IR)~\cite{li2022embracing}.
For the tasks like entity linking~\cite{hoffart2011robust} and knowledge graph canonization~\cite{galarraga2014canonicalizing}, the core challenge lies in modeling the semantic similarity of entities in complex contexts, where understanding the synonymous relationship between phrases is essential.

\begin{figure*}[!t]
    \centering
    \includegraphics[width=0.8\linewidth]{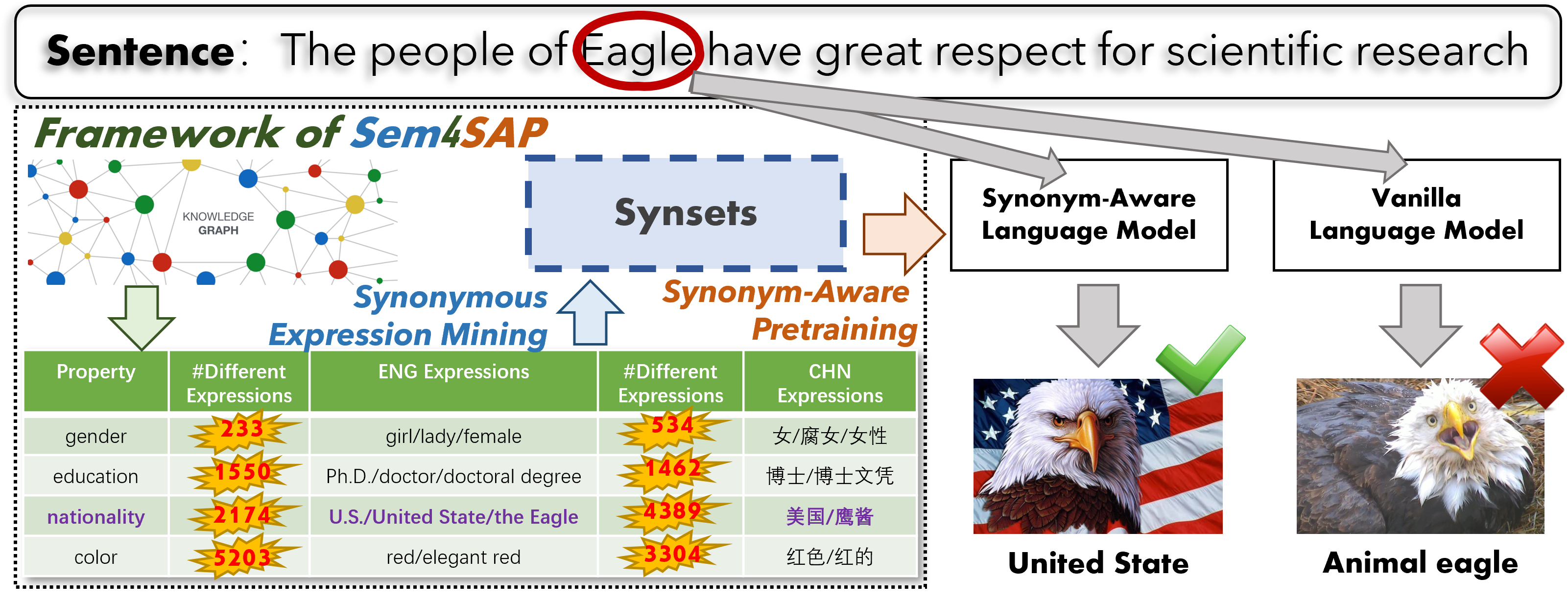}
    \vspace{-2mm}
    \caption{An example of how Sem4SAP works for synonymous expression mining in Open-KG and improving model's understanding of synonymous expressions.
    }
    \label{fig:intro}
    \vspace{-5mm}
\end{figure*}
However, most prevailing language models~(LMs) have limited ability to understand synonymous expressions.
Statistical language models~\cite{cavnar1994ngram} is based on a probabilistic approach where they assign probabilities to each word in a sentence given the context. 
{Without directly using synonymous expressions to do the augmentation, statistical language models will take synonymous expressions only as common words with no similar semantics. }
Pretrained Deep Language Models~(PLMs)~\cite{peters2018deep} play an important role in NLP nowadays, however, they are mainly trained by mask prediction task~\cite{devlin2018bert}, which primarily concern with the probability of a certain sequence of words, but do not concern with the nuances of the word's semantics. 
As a result, PLMs, even large-scale PLMs, tend only to pick a word that fits the context best but not understand the synonymous relation between words.

Based on the above limitations of the existing models, we propose to improve machine's understanding to synonymous expressions from the following two aspects:
\begin{enumerate}
    \item \textbf{Construct large-scale explicit synsets.} It is difficult for the models themself to capture implicit synonymous relations directly from context, let alone enhance the model's understanding of synonymous expressions.
    Starting from explicit synset, which is the set of synonymous expressions, is more feasible at present.
    \item \textbf{Develop synonym-aware pretraining tasks.}
    The prevailing pretraining tasks prevent PLMs from being aware of the synonymous expressions.
    Therefore, novel pretraining tasks which directly inject synonym knowledge into LMs need to be developed.
\end{enumerate}

Although there have some explicit synsets, such as Wordnet~\cite{miller1995wordnet} or BigCilin~\cite{liu2022bigcilin}, it is difficult to directly apply them to facilitate downstream tasks for their homogeneous expression and the limited number of synonymous relation.
We find it feasible to use Open Knowledge Graph~(Open-KG)~\cite{lehmann2015dbpedia} to improve the diversity of the expressions and to enlarge the number of synonymous relations of existing explicit synsets. 
The Open-KG is constructed from the online encyclopedia, which is contributed by a large number of netizens' crowdsourcing.
From the table listed in Figure~\ref{fig:intro}, it is intuitive to find that there are many diverse synonymous expressions in Open-KG.
For example, gender is generally be divided into two categories,
but there have 233 different expressions of gender in Wiki-data~\cite{vrandevcic2014wikidata}, which is one of the most famous Open-KG in the world,
and even 534 different expressions of gender in CN-DBpedia~\cite{xu2017cn}, which is the largest Chinese Open-KG.

The limitation of the existing pretraining tasks lies in lacking of the objectives for learning to represent the relation among synonymous expressions. 
The challenge is that although synonymous expressions have the same semantics most of time, they also have some specific meaning in some sepcific aspects.
So, the representations of two sentences with only synonymous differences should be similar, but the ambiguity should also be kept.

In this paper, we propose \textbf{Sem4SAP}.
As shown in Figure~\ref{fig:intro}, Sem4SAP is a framework of \textbf{S}ynonymous \textbf{E}xpressions \textbf{M}ining in Open-KG \textbf{for} \textbf{S}ynonym-\textbf{A}ware \textbf{P}retraining.
In Synonymous Expression Mining, Sem4SAP leverages the frequency information in Open-KG to detect the core semantic of a expression and uses both textual and distributed representational features of the core semantic to calculate the similarity between each expression for better clustering expressions into synsets.
Besides, by keeping track of the updated content in Open-KG, synsets will always contain the newly emerged expressions, which ensures the output of Sem4SAP is always closely related to the up-to-date Internet. 
In Synonym-Aware Pretraining, Sem4SAP consists of two complementary pretraining tasks, which inject synonymous expression knowledge and synonymous context knowledge into models, and two tricks to keep the specific ambiguity of the synonymous expressions.

In summary, we conclude that this paper makes the following contributions:
\begin{enumerate}
    \item We propose to acquire the up-to-date synsets with greatly diverse synonymous expressions by using the crowd-sourcing content in continually updating Open-KG. 
    \item We propose a novel synonymous expressions mining method to cluster the content in Open-KG into synsets, and we also propose two novel pretraining tasks to inject the synonym knowledge into LMs.
    \item Extensive experiments have been carried out to verify that Sem4SAP outperforms all the other baselines in improving the synonymous expressions understanding to models, and every component in Sem4SAP is effective.
\end{enumerate}

\section{Preliminary}
\label{02}
\subsubsection{Entity, Predicate, Triple, Subject, Object,\\ Knowledge Graph.}  
An \textbf{\emph{entity}} can be a thing, person, place, e.g. \texttt{Apple Inc.}, \texttt{Donald Trump}, \texttt{U.S.}, etc. 
A \textbf{\emph{predicate}} describes the characteristics of the entity. 
A \textbf{\emph{triple}} $<s, p, o>$ is the fact that describes \emph{\textbf{subject}} $s$, while $p$ is a predicate, and $o$ is \emph{\textbf{object}}.
For example, triple $<$\texttt{U.S.}, \texttt{President}, \texttt{Donald Trump}$>$ represents the fact that the president of the U.S. is Donald Trump. 
Then a \textbf{\emph{Knowledge Graph~(KG)}} $G=<E, T>$ consists of entities $E$ and triples $T$. 
Specifically, entities and triples constitute the essential elements of a KG~\cite{lehmann2015dbpedia}.
And Open-KG, which differs from the Knowledge Bases~(KB), is the most primitive KG, while KB is a strictly processed KG.

\subsubsection{Relation, Property, Value.}

Predicate can be further divided into relation and property.
A \emph{\textbf{relation}} is generally used to represent the relationship between entities, while a \emph{\textbf{property}} is used to describe some features of a subject.
\emph{\textbf{Value}} is a concrete description to the subject.
So both entity and value can all be the object to a triple, while only entity can be the subject to a triple.
A relation connects two entities, while a property connects an entity and a value.

\subsubsection{Synonym, Synonymous Expression, Synset.} 
\emph{\textbf{Synonym}} is a word or phrase that means exactly or nearly the same as another word or phrase~\cite{stanojevic2009cognitive}.  
For example, “Aspirin” and “Acetylsalicylic Acid” refer to the same drug; “United States” and “USA” represent the same country. 
\emph{\textbf{Synonymous expression}} is a relaxed form of synonym, which is a sequence of characters that means exactly or nearly the same as another sequence of characters in the same language. 
For example, “gender male” and “man gender” refer to the same gender.
And \emph{\textbf{synset}} is a set of synonyms or synonymous expressions that are grouped together. 

\begin{figure*}[!t]
    \centering
    \includegraphics[width=0.8\linewidth]{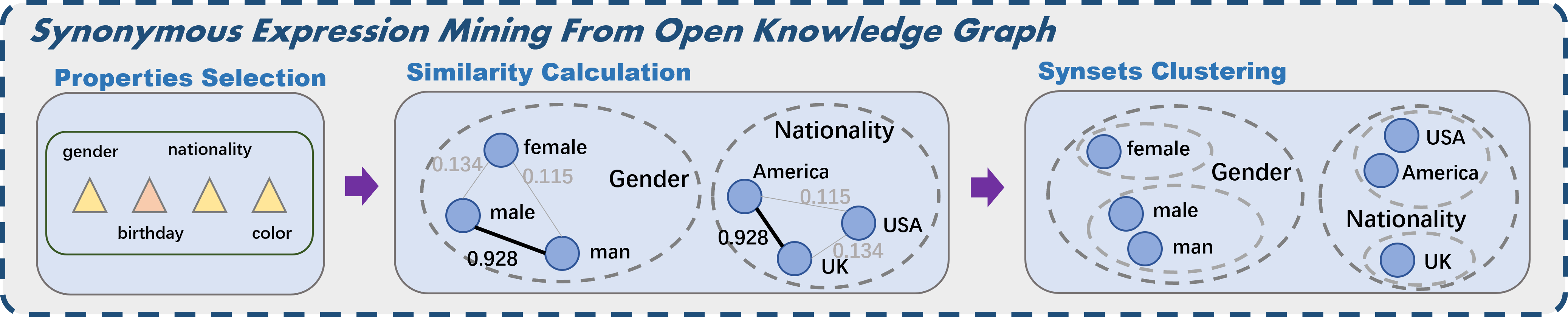}
        \includegraphics[width=0.8\linewidth]{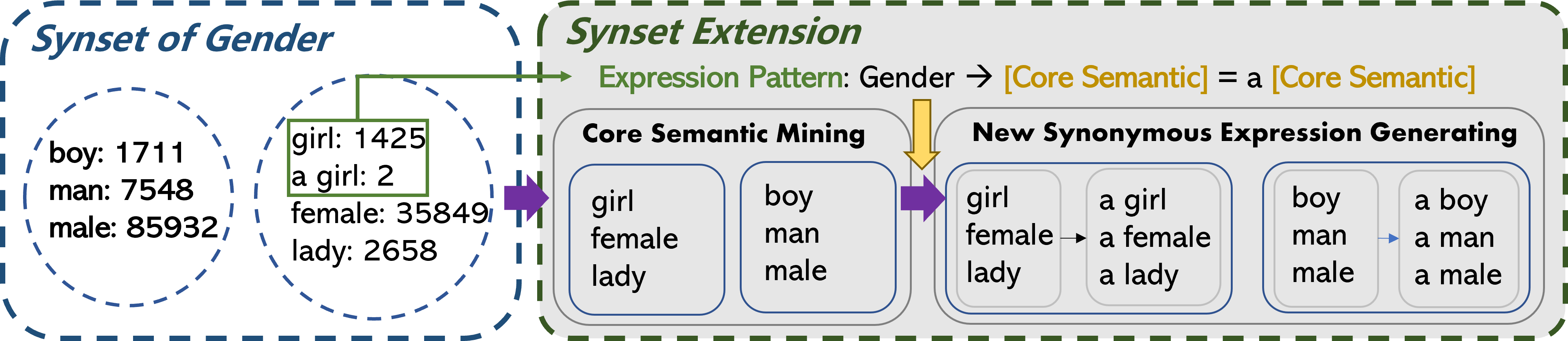}
    \includegraphics[width=0.8\linewidth]{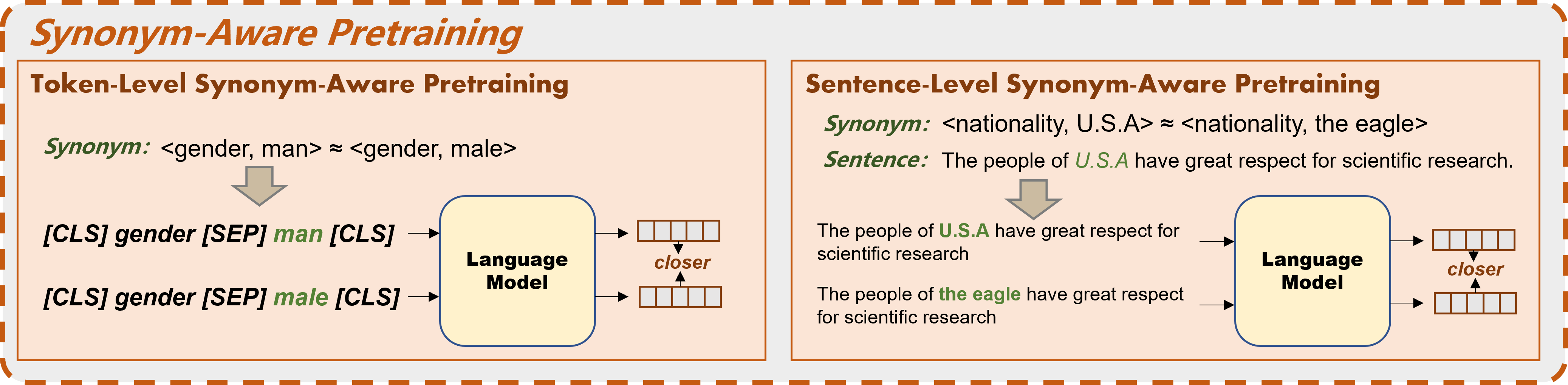}
    \vspace{-2mm}
    \caption{Sem4SAP consists of two main components and a optional component. First, Sem4SAP conduct the Synonymous Values mining in Open-KG. Then Sem4SAP use the mined synsets to do the expansion. Finally, Sem4SAP use two synonym-aware pretraining methods to boost the PLMs' understanding to synonym.}
    \label{fig:my_label}
    \vspace{-5mm}
\end{figure*}
\section{Synonymous Values Mining}
\label{03}
In this section, we elaborate the Synonymous Expressions Mining method of Sem4SAP.
Our method targets the value objects in Open-KG.
Since value object is connected by property but not relation, we detailed our method of distinguishing property from relation in subsection~\ref{021}.
The similarity calculation method is in subsection~\ref{022}.
And we introduce the synsets clustering method in subsection~\ref{023}
Illustration of methods in this section is shown in the upper of Figure~\ref{fig:my_label}

\subsection{Property Selection}
\label{021}
\begin{figure}[!t]
    \centering
    \includegraphics[width=0.85\linewidth]{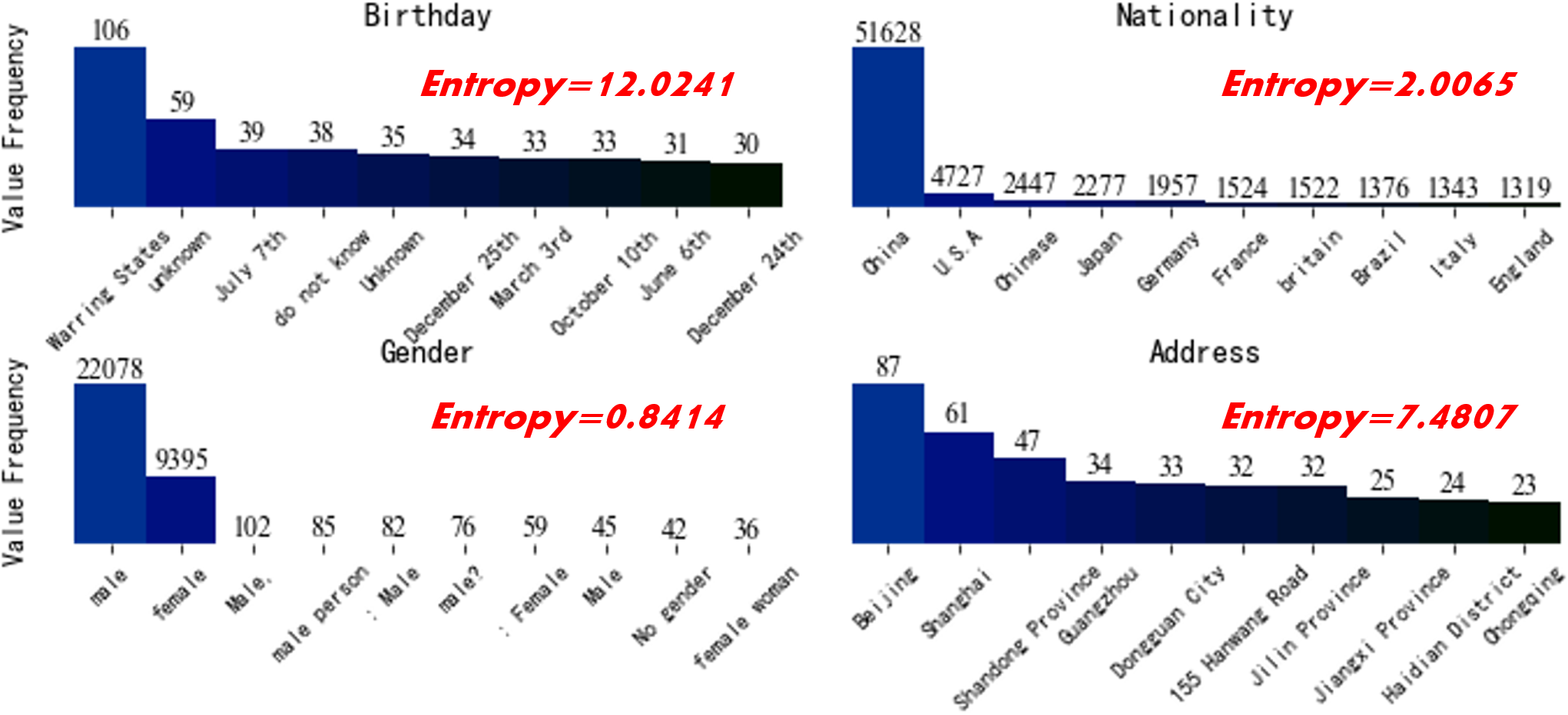}
    \vspace{-2mm}
    \caption{The distribution of objects connected by some predicates and the entropy of these distributions.
    }
    \label{fig:ps.png}
    \vspace{-5mm}
\end{figure}

Since entity contains only specific semantics, while value contains more general semantics, which is expected to facilitate more aspects than entity.
Value only appears as the object in triplet \emph{$\langle$s,p,o$\rangle$}, so the mining targets at the tail of the triplet.
However, predicate is a general semantic constraint to the objects, we expect to find the predicate which has the following characteristics to better find the most descriptive values:
First, predicate should describe some general features of an entity, or to say it should be a property but not a relation.
Second, predicate should describe a categorical feature of an entity but not quantitative.
The distribution of objects has been used to distinguish whether the given predicate is a property of relation~\cite{li2021towards}.
As shown in Figure~\ref{fig:ps.png}, bigger the Shannon entropy of the object distribution is, more possible the predicate is a relation (predicate = \texttt{Address}) or quantitive property (predicate = \texttt{Birthday}), otherwise it is more likely a property (predicate = \texttt{Gender} or \texttt{Nationality}).

Besides, the precision of finding categorical properties can be further improved.
We observe that the value of categorical property tends to have concentrated semantics, like the semantics of gender, which tend to concentrate on the ``male'' or ``female''.
We propose to use the Shannon entropy of the distribution to all possible word-pieces~\cite{wu2016google} (for example, all possible word-pieces to expression ``man'' are ``m'', ``a'', ``n'', ``ma'', ``an'', ``man'') as well as the entropy of the distribution to all objects to distinguish property.
And the formula for calculating the Possibility of Categorical Properties (PCP) is as follows:

\begin{equation}
    \label{equ: s_p}
    \small
    PCP(p_i) = \frac{\# c}{ent(Val(p_i)) * ent(Str(p_i))}\ \forall c\in C_{p_i}.
\end{equation}

$Val(p_i)$ and $Str(p_i)$ represent the value frequency distribution and word-pieces frequency distribution of given predicate $p_i$ respectively.
$C_{p_i}$ denote all characters in the objects connected by the given predicate $p_i$, and $\#c$ denote the number of given character $c$ in all objects of the given predicate, which is used as a penalty term in the formula.
And $ent(p_i)$ represents the Shannon Entropy of the value distribution of $p_i$.

We use Formula~\ref{equ: s_p} to get possibilities for all predicates.
Higher the $PCP(p_i)$ is, more possible $p_i$ is categorical.

\subsection{Similarity Calculation}
\label{022}

We propose to use the frequency information of every word-pieces to detect the important part of the value objects.
Then use textual features and distributed representation features of these important parts to cluster value objects into synsets.

For textual features, we count the number of occurrences of each word-piece of all value objects as the frequency.
And use the frequencies of every word-pieces as the weight of importance.
If two value objects have more high-weighted word-pieces in common, they will share a high similarity for their important part is similar.
Specifically, we use the formula of TF-IDF~\cite{ramos2003tfidf} improved with the frequency of every character to calculate the similarity of two value objects:
\begin{equation}
    \label{TF-IDF}
    TS_{\mathbb{TF-IDF}}(o_m, o_n) = \frac{f(o_m\cap o_n;O_p)}{f(o_m\cup o_n;O_p)},
\end{equation}
where $\mathbb{TF-IDF}$ denote we are using the TF-IDF method to calculate the Textual Similarity,
$O_p$ denote all the value object $o$ of property $p$, and $f(o;O)$ denote the sum of the frequencies from characters $o$ in $O$.

For distributed representation similarity, we mainly use the LM embedding as the distributed representation of the value objects.
Since LMs have limited expressive ability in representing sequences with few tokens~\cite{batista2018language}, we input all word-pieces of a value object into LMs.
The frequencies of each word-pieces of a value object are used as the weights to calculate the representation, which means the important word-piece of a value object will contribute more to the embedding of a value object.
We use cosine similarity to calculate the distributed representation similarity between two value objects, the specific formula is as follows:
\begin{equation}
    \label{LM-embedding}
    \small
    DRS_{\mathbb{M}_j}(o_m, o_n) = \frac{LM_{\mathbb{M}_j}(o_m;\Theta)\cdot LM_{\mathbb{M}_j}(o_n;\Theta)}{\parallel LM_{\mathbb{M}_j}(o_m;\Theta) \parallel\cdot \parallel LM_{\mathbb{M}_j}(o_n;\Theta) \parallel},
\end{equation}
where $LM$ denote the LM model, $\mathbb{M}_j$ denotes the different LM type, and $\Theta$ denotes the different parameters of LM.
$LM_{\mathbb{M}_j}(o_m;\Theta)$

Finally, we combine the above two formulas as follows to get the final Semantic Similarity of two value objects:
\begin{equation}
\label{o-similarity}
\small
\begin{aligned}
SS(&\langle p_i, o_m\rangle,\langle p_j, o_n\rangle)\\&=\begin{cases}0&p_i\neq p_j\\ \prod^{N_{TS}}_{i}TS_{\mathbb{S}_i}(o_m, o_n)*\prod^{N_{DRS}}_{j}DRS_{\mathbb{M}_j}(o_m,o_n)&p_i=p_j\end{cases}.
\end{aligned}
\vspace{-2mm}
\end{equation}

\subsection{Synsets Clustering}
\label{023}
A weighted complete similarity network can be formed after calculating the similarity between all value objects.
The external synonym database can be integrated by setting a new node and adjusting the edge weight in the network.
Some community detection methods can be used to cluster value objects into synsets based on the similarity network.
And in this paper, we propose to remove edges between value objects which have a low similarity between each other, and simply apply Louvain algorithm~\cite{blondel2008fast} to do the clustering.

\begin{table*}
    \small
    \centering
    \resizebox{0.8\textwidth}{!}{
    \begin{tabular}{|c|c|c|c|c|c|c|c|c|c|c|c|}
        \hline
        \multirow{2}{*}{Model} & \multicolumn{5}{c|}{GLUE} & \multicolumn{5}{c|}{CLUE}\\
        & \textbf{SST-2} & \textbf{MRPC} & \textbf{QQP} & \textbf{MNLI-mm} & \textbf{RTE} & \textbf{AFQMC} & \textbf{TNEWS} & \textbf{IFLYTEK} & \textbf{CMNLI} & \textbf{CSL}\\
        \hline
        Bert-base & \cellcolor{lime}89.36 & 85.14 & 67.33 & \cellcolor{green}\textbf{83.10} & 62.85 & \cellcolor{lime}71.59 & 54.74 & 56.48 & \cellcolor{green}\textbf{78.43} & 78.73\\

        \cellcolor{yellow} -BT & \cellcolor{yellow} 93.5 & \cellcolor{yellow} 88.9 & \cellcolor{yellow} 71.2 & \cellcolor{yellow} 83.4 & \cellcolor{yellow} 66.4 & \cellcolor{yellow} 73.70 & \cellcolor{yellow} 56.58 & \cellcolor{yellow} 60.29 & \cellcolor{yellow} 79.69 & \cellcolor{yellow} 80.36 \\
        
        +EDA &89.14&84.16&65.68&80.47&62.71&70.17&52.42&55.14&75.65&76.74\\
        
        +LIBERT &86.48&\cellcolor{lime}85.29$\uparrow$&\cellcolor{lime}69.71$\uparrow$&81.12&\cellcolor{lime}64.51$\uparrow$&70.29&53.27&\cellcolor{lime}56.74$\uparrow$&74.33&77.49\\
        
        
        +PICSO &88.47&84.69&\cellcolor{green}\textbf{70.34}$\uparrow$&82.11&63.56$\uparrow$&71.43&\cellcolor{lime}55.48$\uparrow$&55.40&74.19&\cellcolor{lime}79.16$\uparrow$\\
        
        +Sem4SAP $\uparrow$ & \cellcolor{green}\textbf{92.89}$\uparrow$ & \cellcolor{green}\textbf{88.41}$\uparrow$ & 69.15$\uparrow$ & \cellcolor{lime}83.08 & \cellcolor{green}\textbf{65.16}$\uparrow$ &
        \cellcolor{green}\textbf{73.54}$\uparrow$ & \cellcolor{green}\textbf{\underline{\underline{56.82}}}$\uparrow$ & \cellcolor{green}\textbf{\underline{\underline{60.38}}}$\uparrow$ & \cellcolor{lime}78.41 & \cellcolor{green}\textbf{\underline{\underline{81.22}}}$\uparrow$\\

        \hline
        Roberta-base & \cellcolor{lime}92.36 & \cellcolor{lime}86.41 & 68.45 & \cellcolor{lime}84.78 & \cellcolor{lime}64.37 & \cellcolor{lime}70.87 & 55.53 & 56.52 & \cellcolor{lime}79.45 & 78.20\\

        \cellcolor{yellow} -BT & \cellcolor{yellow} 94.6 & \cellcolor{yellow} 89.8 & \cellcolor{yellow} 71.5 & \cellcolor{yellow} 85.2 & \cellcolor{yellow} 67.3 & \cellcolor{yellow} 74.45 & \cellcolor{yellow} 57.12 & \cellcolor{yellow} 60.88 & \cellcolor{yellow} 80.17 & \cellcolor{yellow} 80.59 \\
        
        +EDA &91.78&86.04&67.14&83.45&62.51&69.43&53.48&55.67&76.64&76.31\\
        
        +LIBERT &86.91&86.13&69.38$\uparrow$&82.47&63.35&70.05&53.44&\cellcolor{lime}56.91$\uparrow$&74.46&74.84\\
        
        
        +PICSO &91.56&85.43&\cellcolor{green}\textbf{72.22}$\uparrow$&82.35&62.19&68.51&\cellcolor{lime}56.41$\uparrow$&56.17&75.43&\cellcolor{lime}79.14$\uparrow$\\
        
        +Sem4SAP& \cellcolor{green}\textbf{93.13}$\uparrow$ & \cellcolor{green}\textbf{87.17}$\uparrow$ & \cellcolor{lime}71.48$\uparrow$ & \cellcolor{green}\textbf{84.96}$\uparrow$ & \cellcolor{green}\textbf{66.35}$\uparrow$ &
        \cellcolor{green}\textbf{73.51}$\uparrow$ & \cellcolor{green}\textbf{\underline{\underline{57.34}}}$\uparrow$ & \cellcolor{green}\textbf{60.70}$\uparrow$ & \cellcolor{green}\textbf{\underline{\underline{80.35}}}$\uparrow$ & \cellcolor{green}\textbf{\underline{\underline{81.28}}}$\uparrow$\\

        \hline
        Roberta-large & 93.17 & \cellcolor{lime}90.54 & 87.60 &\cellcolor{green}\textbf{ 89.45} & \cellcolor{lime}86.34 &72.84 & 57.15 & \cellcolor{lime}59.79 & \cellcolor{green}\textbf{81.11} & 80.10\\

        \cellcolor{yellow} -BT & \cellcolor{yellow} 96.7 & \cellcolor{yellow} 92.3 & \cellcolor{yellow} 90.2 & \cellcolor{yellow} 90.2 & \cellcolor{yellow} 88.2 & \cellcolor{yellow} 74.02 & \cellcolor{yellow} 57.86 & \cellcolor{yellow} 62.55 & \cellcolor{yellow} 81.70 & \cellcolor{yellow} 81.36 \\
        
        +EDA &\cellcolor{lime}93.25$\uparrow$&88.37&86.51&\cellcolor{lime}89.22&85.48&\cellcolor{lime}73.10$\uparrow$&55.42&59.64&80.32&78.61\\
        
        +LIBERT &89.64&87.14&85.79&86.97&85.29&71.26&56.43&58.32&78.44&79.71\\
        
        
        +PICSO $\uparrow$&92.48&90.04&89.17$\uparrow$&83.56&85.46&72.43&\cellcolor{lime}57.32$\uparrow$&58.25&76.42&\cellcolor{lime}80.56$\uparrow$\\
        
        +Sem4SAP & \cellcolor{green}\textbf{96.14}$\uparrow$ & \cellcolor{green}\textbf{91.47}$\uparrow$ & \cellcolor{green}\textbf{89.02}$\uparrow$ & 89.17 & \cellcolor{green}\textbf{87.77}$\uparrow$ & 
        \cellcolor{green}\textbf{74.25}$\uparrow$ & \cellcolor{green}\textbf{\underline{\underline{57.64}}}$\uparrow$ & \cellcolor{green}\textbf{62.46}$\uparrow$ & \cellcolor{lime}81.04 & \cellcolor{green}\textbf{\underline{\underline{82.41}}}$\uparrow$\\
        \hline
    \end{tabular}
    }
    \vspace{-2mm}
    \caption{Overall Experiments on GLUE and CLUE datasets}
    \label{tab:overallExperiemnt}
    
    \vspace{-5mm}
\end{table*}
\section{Synsets Expansion}
\label{04}
In this section, we propose an extension method towards adding redundancy to the synsets even though the mined synsets are enough for doing synonym-aware pretraining.
Since there has much redundant information in synonymous expressions, understanding synonymous expression is more difficult than synonyms.
Adding redundant information to the synsets will help models identify the irrelevant information of the expressions, which will improve the robustness of the models.


By adding redundancy, we aim to find the expression pattern, which is composed of important and unimportant parts to deliver core semantic or auxiliary semantics.
We propose to use frequency $f$, Point-wise Mutual Information (PMI)~\cite{church-hanks-1990pmi}, and left-right neighborhood characters richness of every word-piece to find its core part.

PMI is a measurement for the association of two given characters, referring to the probability of a combination of characters occurring as a whole word.
Left and right neighborhood characters' richness is also a metric to detect whether the given sequence is a complete word or not, and it is often calculated by the entropy of their left and right neighboring characters.
The higher the entropy of the left and right neighboring characters is, the richer the left and right neighborhood characters' richness is, which means the word-piece is more likely to be used as a complete word.

We propose the following formula to calculate the Probability of Core Semantic (PCS) of every word-pieces of an expression:
\begin{equation}
\small
    PCS(x,y) = f_{xy} * PMI(x,y) * lrEnt(xy).
\end{equation}
$f_{xy}$ denote the importance of word-piece $xy$, which is concatenate by word-piece $x$ and $y$.
$PMI$ and $lrEnt$ denote PMI metric and left-right entropy.

We choose word-pieces with higher $PCS$ scores to do the replacement with each other within the same property just like the example in the middle of Figure~\ref{fig:my_label}.
In this way, we will get expanded synsets.



\section{Synonym-Aware Pretraining}
\label{05}

In order to make better use of the mined synsets, we design a pretraining method for synonym-aware LMs.
The pretraining method consists of two pretraining tasks to inject synonym knowledge into models, and two extra tricks to prevent overfitting and semantic drifting of the models.

\textbf{Token-level Synonym Boosting:} 
We first design a token-level synonym boosting task for injecting vocabulary synonym knowledge into LMs.
This task allows the LM to consider two synonymous expressions as having the same semantics.
So the objective of this task is to pull in the output of two expressions if they are synonymous to each other, otherwise, remain unchanged.

\textbf{Sentence-level Synonym Boosting:}
Then we design a sentence-level synonym boosting task for injecting contextual synonym knowledge into LMs.
This task allows the LM to consider two sentences as having the same semantics if there is only a synonym substitution difference between them.
And the objective of this task is just like the previous that pull in the output of two sentences if they are synonymous.

\textbf{Two extra tricks:}
We design two extra tricks to prevent overfitting and semantic drifting, especially for continual pretraining based on other PLMs.

The first trick is to fix the output of the most common expressions and pull in the outputs of the others in the token-level synonym pretraining task.
Since the most common expressions are trained most often, their representations are more expressive than the other expressions.
In this way, the semantics of these most expressive expressions are delivered to the other synonymous expressions.

The second trick is to stop pulling in when the distance between the outputs of synonymous expressions is much shorter than the initial distance in both synonym boosting pretraining tasks.
It ensures that the model's comprehension of similar texts will not become identical, which remains the ambiguity of different expressions in some aspects.

\begin{figure*}
    \centering
    \resizebox{0.85\textwidth}{!}{
    \includegraphics{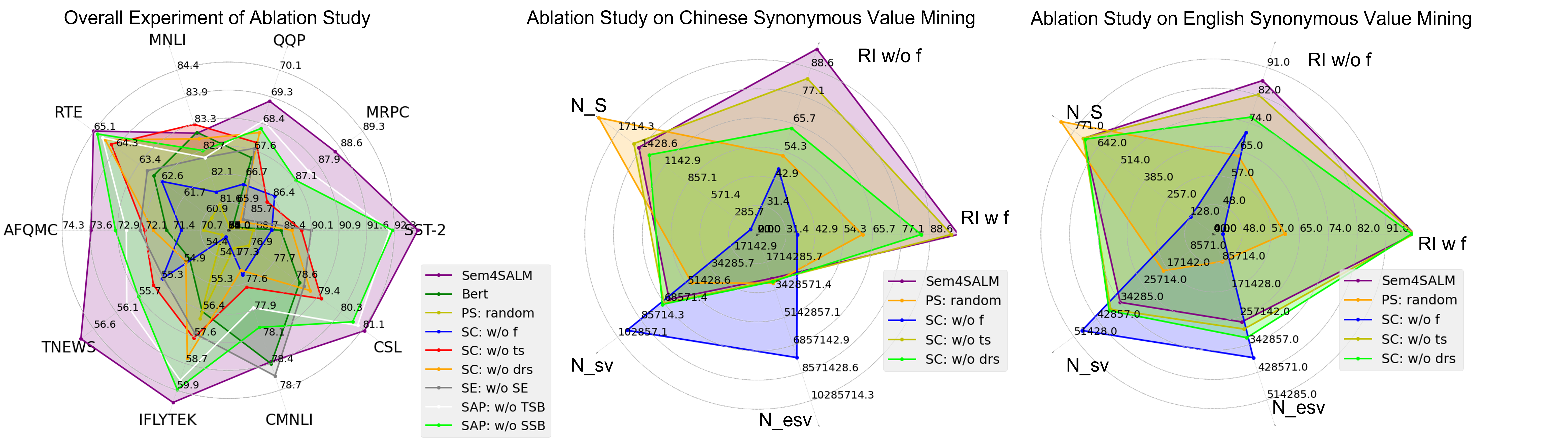}
    }
    \caption{The ablation study of our synonym mining method. The left chart shows the final results of downstream tasks. Middle's shows the result on Synset Mining on DBpedia. Right's shows the result on Synset Mining on CN-DBpedia.}
    \label{fig:ablation}
    \vspace{-5mm}
\end{figure*}
\section{Experiments}
\label{06}
In this section, we evaluate the effectiveness of Sem4SAP.
The setup of our experiments is detailed in Sec.~\ref{061}.
The overall experiments of Sem4SAP and other baselines are detailed in Sec.~\ref{062}.
The ablation study is shown in Sec.~\ref{063}.
And the case study is in Sec.~\ref{064}.

\begin{table}
    \small
    \centering
    \resizebox{0.8\columnwidth}{!}{
    \begin{tabular}{c|c|c|c}
        \hline
        \textbf{Word}& \textbf{ENG Synset} & \textbf{CHN Synset} & \textbf{WordNet 3.0}\\
        \hline
        Female & 11 & 16 & 3 \\
        U.S. & 7 & 8 & 11 \\
        Ph.D. & 10 & 12 & 2 \\
        ♂ & 14 & 22 & 0 \\
        \hline
        \emph{\#Words} & 34,329 & 65,010 & 147,306 \\
        \emph{\#Expended} & 272,745 & 2,719,864 & - \\
        \hline
    \end{tabular}
    }
    \vspace{-2mm}
    \caption{Results and examples of synsets mined by Sem4SAP. Second row and third row denote the number of synonymous expressions to the word in first row in our mined synsets, and fourth row denote the number in Wordnet.}
    \vspace{-5mm}
    \label{tab:synsets}
\end{table}

\subsection{Experiments Setup}
\label{061}

\subsubsection{Datasets}
We conduct synonymous expressions mining in English Open-KG Dbpedia~\cite{lehmann2015dbpedia} and Chinese Open-KG CN-DBpedia~\cite{xu2017cn}, the result and some examples of our mined synsets have listed in Table~\ref{tab:synsets}.
Since these two Open-KGs are composed of content in the general area, so we use General Language Understanding Evaluation benchmark GLUE~\cite{wang2018glue}, and Chinese general Language Understanding Evaluation benchmark CLUE~\cite{xu2020clue} to verify the effectiveness of Sem4SAP.
These benchmarks consist of many datasets, and we choose ten of them that we think best suit our evaluation, the detailed description, and statistics will be listed in the Appendix.
Besides, since we aim at improving the model's understanding of synonymous expressions, so we add back translated data to the test set of these datasets.
Specifically, we use Youdao Translation API
to generate new data by translating the data in the test set into English or Chinese and back-translating them, and adding new data to the test set.
Back translation will bring in many synonymous expressions, which is able to verify whether the models will be affected by synonymous substitution attacks.

\subsubsection{Baselines}
We choose several works that can also enhance PLM's understanding of synonymous expressions as baselines.
We also choose Bert and Roberta as the backbone PLMs.
The abbreviation and description of these works are as follows:

\textbf{Bert}~\cite{devlin2018bert} is one of the most famous PLMs which shows its great ability in natural language understanding.
\textbf{Roberta}~\cite{liu2019roberta} has made progress in the pretraining technique which performs better than Bert in natural language understanding.
\textbf{-BT} denotes the performance of the original dataset.
\textbf{EDA}~\cite{wei2019eda} is one of the most famous text data augmentation methods in using four text transformations, which are synonym substitution, random insertion, random swapping, and random deletion, to enhance the performance of the model. 
\textbf{LIBERT}~\cite{lauscher2019specializing} trains Bert from scratch with an auxiliary task that binary classifies whether entity pairs are synonymous pairs.
\textbf{PICSO}~\cite{li2022embracing} use an adapter to store the synonymous knowledge without undermining the raw knowledge in PLM.
And \textbf{Sem4SAP} is our proposed method.

\subsubsection{Parameter Setting}

In synonymous expressions mining phrase, we choose 5000 properties in Chinese and English Open-KGs respectively with the lowest $PCP$ score, and clear 40\% edges with the lowest weight during the clustering.
In Synonym-Aware Pretraining, we take the expressions that match exactly the value objects of the highest frequency in Open-KG as the most common expressions.
We also choose to stop pulling in the output of the synonymous expressions or sentences when their distance has already been pulled in at least 60 \% shorter than the initial distance.


\subsubsection{Fairness of the Experiments}
To keep the fairness of the experiments, we first acquire the text from the training data in all datasets listed in Table~\ref{tab:overallExperiemnt}.
Then, we conduct the synonym substitution to the acquired text by replacing the words or phrases in the text which match exactly to the expressions in mined synsets.
Finally, we have at least pretrain all baselines in the way how backbone PLMs are trained.
So, the pretraining corpus for all backbone PLMs is the same, the only difference is the pretraining method.

\subsection{Overall Experiments}
\label{062}
We compare baselines on both GLUE and CLUE dataset, and the experimental results are listed in Table~\ref{tab:overallExperiemnt}.
The \emph{yellow} cell denotes the result of the original dataset without doing back translation, and most of these results are cited from the paper of Robert, and Bert and from the leaderboard of GLUE and CLUE.
The \emph{green} cell denotes the best result, and the \emph{lime} cell denotes the second-best result.
The $\uparrow$ denotes whether the performance is higher than the raw PLM's.
The \underline{\underline{double underline}} denotes whether the performance on the back translation dataset is even higher than the performance on the original dataset.

\textbf{Robustness of PLMs:} From the result, we can see that the performances of the PLMs on original datasets are mostly better than the performances on the dataset added back to translation data, which represents that the existing PLMs are all suffering from substitution attacks.

\textbf{Performance of Baselines:} The EDA and LIBERT mostly lower the performance of the raw PLM, this is because they introduce synonym knowledge improperly, which impairs the existing knowledge in PLMs.
PICSO perform well on QQP, TNEWS, and CSL datasets, which correspond to the results in their paper.

\textbf{Performance of Sem4SAP: }Sem4SAP out-perform most other baselines in most tasks.
Sem4SAP is also a stable method that achieves better performance in all the tasks than raw PLM except for some Natural Language Inference tasks (MNLI-mm, CMNLI).

\textbf{Performance of Large-scale PLM: }For models with larger parameters (Roberta-large), most baselines only inject harmful information to the model while ours further improve the performance.
Moreover, for large models, Sem4SAP still makes the performance better than some of the original datasets, which means using Sem4SAP to inject synonym knowledge is also an effective way of text data augmentation to the large-scale PLMs.

\subsection{Ablation Study}
\label{063}
The result of ablation studies is presented in Figure~\ref{fig:ablation}.
The left figure shows the performances on different downstream applications by removing seven components in Sem4SAP respectively.
The middle and right figures show the quality of mined synsets in Chinese and English language by removing four components of our synonym mining method one by one, the results are annotated by three graduate students.
The label in the Figure~\ref{fig:ablation} are listed as follows:
``\textbf{PS: random}'' denote we randomly choose the properties in Property Selection phrase.
``\textbf{SC: w/o f}'' denote we don't use the frequency information in Synset Clustering phrase.
``\textbf{SC: w/o ts}'' denote we remove the textual similarity in Synset Clustering phrase.
``\textbf{SC: w/o drs}'' denote we remove the distributed representational similarity in Synset Clustering phrase.
``\textbf{SE: w/o SE}'' denote we don't do the Synset Expansion.
``\textbf{SAP: w/o TSB}'' denote we remove the Token-level Synonym Boosting method in Synonym-Aware Pretraining phrase.
``\textbf{SAP: w/o SSB}'' denote we remove the Sentence-level Synonym Boosting method in Synonym-Aware Pretraining phrase.

We use Bert-base model in all three figures.
In the left figure, we use accuracy metric to evaluate the result of ablation study.
In the middle and right figure, we use five additional metrics to present the result of the clustering:
``\textbf{N\_S}'' represents the number of Synsets mined by Sem4SAP.
``\textbf{N\_sv}'' represents the number of different value objects mined by Sem4SAP.
``\textbf{N\_esv}'' represents the number of synonymous expressions obtained by Sem4SAP after doing synset expansion.
``\textbf{RI}'' indicate the Rand Index~\cite{rand1971objective}, which is a metric to denote the clustering purification。
``\textbf{w f}'' or ``\textbf{w/o f}'' denote calculating RI with frequency of the value objects or not.

From Figure~\ref{fig:ablation}, we can find that the method of Property Selection and using frequency to do the clustering are the two most important designs of our synonymous expression mining method.
Few synonyms and many synsets are clustered by randomly selecting properties, which means only a few synonyms in each synset, and the clustering purity is really low, resulting in the smallest covering area in the left figure.
When frequency information is not used, the mined synsets will contain a large number of synonyms, but the clustering purity is very low, resulting in the second smallest covering in the left figure.
And Sem4SAP will be better than bert-base when we use at least the above two methods.
Moreover, we found that removing the token-level or sentence-level synonym boosting method will reduce the performance of the model a lot, so we conclude that these two methods are complementary. 
The token-level synonym boosting method enhances the ability of the model to understand specific synonyms, while the sentence-level synonym boosting method enhances the ability of the model to understand synonyms in context.

\subsection{Case Study}
\label{064}
Here we give some cases of Sem4SAP in Table~\ref{tab:case}.
The first table is the cases of our mined synsets.
Although there have some symbols in Open-KG, ``♂'' for example, Sem4SAP still understand its meaning by mining the co-occurrence of other words, like ``男性'' (male) $\rightarrow$ ``男♂'' (male ♂) $\rightarrow$ ``♂的'' (gender of ♂).
Sem4SAP cluster ``美国 (U.S.A)'' and ``鹰国 (Country of Eagle)'' in the same synset, which shows Sem4SAP has a good understanding of slangs.
We think this is because ``美国 (U.S.A)'' and ``老鹰 (Eagle)'' may have some co-occurrence in pre-training data of PLMs, and Sem4SAP can well integrate the existing knowledge in PLMs and also capture the important parts of the words to better understanding them.

The second table gives some cases of how performance changes by using Sem4SAP.
But it still suffers from synonym substitution attacks.
When the Chinese phrase ``找私房钱'' (look for own money) changes into ``找钱'' (look for money, give change), the model will not identify it as a sentence of amusement anymore.
If the English phrase ``end raunchy'' changes into ``end crass'', the model may identify that ``crass'' is a much more negative word than ``raunchy'', then identify the sentence as ``Negative Emotion''.
And Sem4SAP can make models understand that these are synonyms and that they should have similar semantics under the given context, which will improve the performance in downstream tasks.

\begin{table}[!t] 
\small
\centering
\resizebox{0.9\columnwidth}{!}{
\begin{tabular}{|l|}
\hline
    \makecell[c]{\textbf{Mined Synset}} \\
    \hline
    \hdashline
    \textbf{Case 1}~\cmark\\
    \textcolor{blue}{\textbf{Gender}} \\
    男性 (male), 男♂ (male ♂), 男孩子 (boy), ♂的 (of ♂) \\
    \hdashline
    \textbf{Case 2}~\cmark\\
    \textcolor{blue}{\textbf{Status of the Novel}} \\
    暂停 (pause), 持续暂停 (pausing), 正在休息 (resting)\\
    \hdashline
    \textbf{Case 3}~\xmark \\
    \textcolor{blue}{\textbf{Nationality}} \\
    美国 (U.S.A), 鹰国 (Country of Eagle), 鹰酱 (Cute Eagle)\\
    \hline
    \makecell[c]{\textbf{Performance Changes in Downstream Tasks}} \\
    \hline
    \hdashline
    \textbf{Case 1}\\
    \textcolor{blue}{\textbf{Dataset: TNews}} \\
    \textcolor{purple}{\textbf{Label: 休闲益智 (Amusing)}} \\
     \textcolor{c1}{\textbf{Roberta-large}: 自己的私房钱还是得自己来找了。}~\cmark\\
     \textbf{Translate}: I'm gonna have to find my own cash.\\
     \textcolor{c1}{\textbf{Roberta-large}: 我得自己去找钱了。}~\xmark\\
     \textcolor{c2}{\textbf{Sem4SAP: }自己的私房钱还是得自己来找了。}~\cmark\\
     \textbf{Token-level boosting}: 私房钱, 私房钞票, 钱~(money, hidded money)\\
     \textbf{Sentence-level boosting}: 另类解谜游戏，可以教你如何藏钱哦。\\
     \textbf{Sentence-level boosting}: 另类解谜游戏，可以教你如何藏私房钱哦。\\
      (An alternative puzzle game that will teach you how to hide money)\\
    \textcolor{c2}{\textbf{Sem4SAP: }我得自己去找钱了。}~\cmark\\
     
    \hdashline
    \textbf{Case 2}\\
    \textcolor{blue}{\textbf{Dataset: SST}} \\
    \textcolor{purple}{\textbf{Label: Positive Emotion}} \\ 
    \textcolor{c1}{\textbf{Roberta-large}: It's is your ticket right here to end raunchy college humor.}~\cmark\\
    \textbf{Translate}: 这是你的一张票，你将结束粗俗的大学幽默。\\
    \textcolor{c1}{\textbf{Roberta-large}: This is your ticket, you will end crass college humor.}~\xmark\\
    \textcolor{c2}{\textbf{Sem4SAP: }It's your ticket right here to end raunchy college humor.}~\cmark\\
    {\textbf{Token-level boosting}}: raunchy, crass, raunchy life\\
    {\textbf{Sentence-level boosting}}: Crass, then gasp for gas, verbal deportment \\
     \textbf{Sentence-level boosting}: Raunchy, then gasp for gas, verbal deportment \\
     \textcolor{c2}{\textbf{Sem4SAP: }This is your ticket, you will end crass college humor.}~\cmark\\
    \hline
\end{tabular}
}
\vspace{-2mm}
\caption{Case study of our proposed synonym mining method and text data augmentation algorithms}
\label{tab:case}
\vspace{-5mm}
\end{table}

\section{Related Works}

\subsection{Synonym Detection}
Early efforts on synonym detection focus on finding entity synonyms from a semi-supervised situation such as query logs~\cite{ren2015synonym}, web tables~\cite{he2016automatic}, and synonymy dictionaries~\cite{ustalov2017watset}. 
In comparison, our work aims at developing a framework to cluster all chosen vocabulary terms into synsets without any supervised signals.
Modern synonym detection combines entity set expansion task and synonym detection task which are considered tightly coupled to each other~\cite{shen2020synsetexpan}, and the syntactic dependencies are more developed by using deep model to find synonymous meaning~\cite{yang2022synonym}.
There are other works who tries to find synonyms in an Open-KG~\cite{qu2017automatic} especially when there is a need to canonize an Open-KG, but most them utilize side information to compensate for the lack of semantic~\cite{vashishth2018cesi,lin2019canonicalization,shen2022multi}, others detecting synonyms by jointly conducting entity linking~\cite{liu2021joint}.
However, they ignore that not all the content in KG is suitable for synonym detection, so their methods are too time-extensive and results are noisy. 
Our methods first detect which terms are suitable to do the synonym detection, and utilize three heterogeneous features to extract the semantics of the terms and to cluster them into synsets.

\subsection{Synonym Augmentation}
There are many efforts have been made to use synonyms to enhance the effect of downstream applications.
LIBERT~\cite{lauscher2019specializing} trains Bert from scratch with an auxiliary task that binary classifies whether entity pairs are synonymous pairs.
SAPBERT pre-trains BERT with synsets from Unified Medical Language System, which is a comprehensive collection of biomedical terms.
Some choose to use a synonym to construct sentences by simply using templates and conducting Mask-Word-Predication task~\cite{yuan2022generative}. 
And PICSO uses an adapter to store the synonym knowledge to prevent the undermining of the original PLMs~\cite{li2022embracing}.
We aim to develop specific pre-training tasks for injecting synonym knowledge from explicit synsets into LMs. 
The tasks we developed not only ensures the enhancement of PLM's ability to understand synonyms, but also ensures that no semantic drift will be introduced.

\section{Conclusion}
In this paper, we aim to boost the performance of the models in synonym understanding.
We propose the Sem4SAP framework, which consists of three main components.
First, we do the Synonymous Values mining in Open-KG. 
Then we expand the mined synsets. 
Finally, we use two synonym-aware pretraining methods to boost the PLMs’ understanding of synonyms.
In the future, we want to use our method to mine more synsets and release them after manual annotation.



\newpage
\bibliographystyle{named}
\bibliography{ijcai22}


\end{CJK}
\end{document}